\documentclass[colorlinks=true, linkcolor=webgreen, urlcolor=webgreen,  citecolor=webblue]{article}
\usepackage{arxiv}

\usepackage[utf8]{inputenc} 
\usepackage[T1]{fontenc}    
\usepackage{hyperref}       
\usepackage{url}            
\usepackage{booktabs}       
\usepackage{amsfonts}       
\usepackage{nicefrac}       
\usepackage{microtype}      
\usepackage{lipsum}		
\usepackage{graphicx}
\usepackage{natbib}
\usepackage{doi}
\usepackage{authblk}
\usepackage{amsmath}
\usepackage{hyperref}
\usepackage{xcolor}
\definecolor{webgreen}{rgb}{0,0,0}
\definecolor{webbrown}{rgb}{.6,0,0}
\definecolor{webblue}{RGB}{0,0,128}
\title{CLSRIL-23: Cross Lingual Speech Representations for Indic Languages}

\author[1]{Anirudh Gupta\thanks{anirudh.gupta@thoughtworks.com}}
\author[1]{Harveen Singh Chadha\thanks{harveen.chadha@thoughtworks.com}}
\author[1]{Priyanshi Shah}
\author[1]{Neeraj Chhimwal}
\author[1]{Ankur Dhuriya}
\author[1]{Rishabh Gaur}
\author[2]{Vivek Raghavan}
\affil[1]{Thoughtworks}
\affil[2]{Ekstep Foundation}

\date{} 					

\renewcommand{\headeright}{}
\renewcommand{\undertitle}{}
\renewcommand{\shorttitle}{}

\hypersetup{
pdftitle={A template for the arxiv style},
pdfsubject={q-bio.NC, q-bio.QM},
pdfauthor={David S.~Hippocampus, Elias D.~Striatum},
pdfkeywords={First keyword, Second keyword, More},
}

\begin{document}
\maketitle

\begin{abstract}
	We present a CLSRIL$\mbox{-}23$, a self supervised learning based audio pre-trained model which learns cross lingual speech representations from raw audio across $23$ Indic languages. It is built on top of wav2vec $2.0$ which is solved by training a contrastive task over masked latent speech representations and jointly learns the quantization of latents shared across all languages. We compare the language wise loss during pretraining to compare effects of monolingual and multilingual pretraining. Performance on some downstream fine-tuning tasks for speech recognition is also compared and our experiments show that multilingual pretraining outperforms monolingual training, in terms of learning speech representations which encodes phonetic similarity of languages and also in terms of performance on down stream tasks. A decrease of $5$\% is observed in WER and $9.5$\% in CER when a multilingual pretrained model is used for finetuning in Hindi. All the code models are also open sourced. CLSRIL-$23$\footnote{\url{http://github.com/Open-Speech-EkStep/vakyansh-models}} is a model trained on $23$ languages and almost $10,000$ hours of audio data to facilitate research in speech recognition for Indic languages. We hope that new state of the art systems will be created using the self supervised approach, especially for low resources Indic languages. 
\end{abstract}

\section{Introduction}
Speech recognition has made remarkable progress in the past few years especially after the advent of deep learning \citep{hinton}. End to end (E2E) models have surely simplified the modelling process but they are also notoriously known for huge amount of data requirements. Especially more so for low resource languages \citep{besacier2014automatic}.

This is of particular importance for countries with many languages and dialects such as India which has 22 official languages with an additional 1500 minor languages/dialects. Apart from a few major languages, most of the languages are low resource, thereby making it more difficult to develop speech related technologies \citep{bourlard2011current}.

(E2E) networks become an attractive choice for multilingual ASRs since they combine the acoustic model, pronunciation and lexicon model into a single network. One way to tackle with multiple languages using a single model would be to train a multilingual ASR model we can take a union over all the language characters and jointly train a model on all the languages. But even in that approach huge amounts of data is needed per language.

In recent years self supervised learning has emerged as a new paradigm in which representations are learnt from the data itself and then fine tuning is done on several other down stream tasks. This approach has been widely successful in natural language processing (NLP) applications \citep{devlin2019bert,peters-etal-2018-deep} and is active area of research in other fields. 

In the past few years self supervised learning has been actively studied for speech recognition. In \citep{jiang2019improving} the authors perform unsupervised pretraining with masked predictive coding using a transformer model. Most of work in this space as well has been in monolingual speech recognition \citep{chung2018speech2vec,tjandra2019vqvae,jiang2019improving,harwath2020learning}.
Our approach is based on the wav2vec $2.0$ \citep{baevski2020wav2vec} the details of which are explained in the coming sections. An approach which uses multiple languages in pre-training and fine-tuning is described in \citep{conneau2020unsupervised}. It also shows that cross lingual pre-training outperforms monolingual pre-training. We extend the work by \citep{rivire2020unsupervised} and \citep{conneau2020unsupervised} by pretraining only on Indic languages so that speech recognition tasks have a better performance on Indic languages.

Languages spoken in the South Asian region belong to at least four major language
families: Indo-European (most of which belong to its sub-branch Indo-Aryan), Dravidian, Austro-Asiatic, and Sino-Tibetan. Almost one third of our mother-tongues in India ($574$ languages) belong to the Indo-Aryan family of languages - spoken by $73.30$\% of Indians. The Dravidian languages, $153$ in number, form the second major linguistic group of the country ($24.47$\% )\footnote{\url{https://www.education.gov.in}}. Since most of the $23$ we have used are in common language families we aim to utilise language similarity to aid representation learning for low resource languages.

\section{Modeling Approach}

The method we use masks the speech input in the latent space and solves a contrastive task defined over a quantization of the latent representations which are jointly learned \citep{baevski2020wav2vec} , and shows that powerful representations can be learnt from speech audio alone. The approach encodes raw speech audio via a multi layer convolutional network and then masks resulting latent speech representations, similar to \citep{devlin2019bert}. The latent representations are fed to a Transformer network to build contextualized representations and the model is trained via a contrastive task where the true latent is to be distinguished from distractors. The discrete speech units are learnt by a Gumbel softmax \citep{jang2017categorical} to represent the latent representations in the contrastive task. 

\subsection{Pre-training}

The main body of the architecture consists of a CNN based feature encoder, a Transformer based sequence network and a quantization module. The feature encoder maps the raw waveform $X$ to latent speech representations $Z$. These representations are then fed to a transformer block to generate context representations $C$, capturing the information in the entire sequence. The quantization module is used to discretize latent speech representations $Z$ into $Q$. 
Given $G$ codebooks with $V$ entries where dimension of each codebook is $\mathbb{R}^{V \times d/G}$. Row corresponds to an entry in the codebook. We choose one entry from each codebook and concatenate the resulting vectors $e_1, e_2, ..., e_G$. This concatenated vector is then mapped to $q$ using a linear transform. The Gumbel softmax enables choosing discrete codebook entries in a fully differentiable way and probabilities for choosing the $v$-th codebook entry of group $g$ are,

\begin{equation}
p_{g,v} = \dfrac{exp(l_{g,v} + n_v)/\tau}{\sum_{k=1}^{V} exp(l_{g,v} + n_v)/\tau}
\end{equation}

where $\tau$ is a non-negative temperature, $n=-log(-log(u))$ and $u$ are uniform samples from $\mathcal{U}(0,1)$.

The speech representations during pre-training are learnt by a contrastive task $\mathcal{L}_m$. This is augmented by a codebook diversity loss $\mathcal{L}_d$. 

\begin{equation}
\mathcal{L} = \mathcal{L}_m + \alpha \mathcal{L}_d
\label{total-loss}
\end{equation}

where $\alpha$ is a tuned hyperparameter. Where,

\begin{equation} \label{lm}
\mathcal{L}_m = -log \dfrac{exp(sim(q_t, c_t) / \kappa}{\sum_{\widetilde{q} \sim Q_t}exp(sim(\widetilde{q}_{t}, c_t) / \kappa)}
\end{equation}

\begin{equation}
\mathcal{L}_d = \dfrac{1}{GV} \sum_{g=1}^{G} \sum_{v=1}^{V} \bar{p}_{g, v} \log \bar{p}_{g,v}
\end{equation}

$\mathcal{L}_m$ is the contrastive loss to make the model distinguish true representations from latent distractors $\widetilde{q}$. In equation \ref{lm} $sim$ is the cosine similarity. The diversity loss $\mathcal{L}_d$ is designed to increase use of the quantized notebook representations. 

\subsection{Fine-tuning}
Pre trained models are fine-tuned by adding a fully connected layer on top of the context network with the size of output layer equal to the vocabulary of the task. Models are optimized using a CTC loss \citep{Graves06connectionisttemporal}. During training only weights of the transformer module are updated but not of the feature encoder module. 

\section{Training Data}

All our data has been processed through the open sourced framework called \textit{Vakyansh}\footnote{\url{https://open-speech-ekstep.github.io/}}. The basic steps of the process are - 

\begin{itemize}
	\item Download and convert audio to \textit{wav} format with sample rate $16000$, number of channels $1$ and bit rate per sample of $16$. 
	\item We split an audio into voiced chunks using voice activity detection\footnote{\url{https://webrtc.org/}}. We make sure that all the voiced chunks lie between $1$ and $30$ seconds.
	\item To detect and reject noisy samples we use a signal to noise ratio (SNR) approach described by \citep{wadasnr}. We consider any audio sample below a SNR value of $25$ as noise and do not include them in training data.
	\item We perform speaker and gender identification on our audio data. A high level representation of voice is learnt using a voice encoder based on \citep{wan2020generalized}. For each audio sample the voice encoder creates a $256$ dimensional encoding that summarizes characteristics of the spoken voice. For gender identification we train a support vector machine algorithm on the embedding with manually labelled data. Our goal for speaker identification was to get a sense of the number of speakers in a particular audio source. To estimate we use a hierarchical clustering approach to cluster \textit{similar} embeddings in the sense of cosine similarity. The number of speakers are thus the number of clusters. 
\end{itemize}

\begin{table}

	\centering
	\begin{tabular}{llllllllllll}
		\toprule
		                   
		& Assamese  & Bengali & Bodo & Dogri & English & Gujarati & Hindi & Kannada  & \textbf{Total} \\
		\midrule
		\textit{train} & $254.9$  & $331.3$ & $26.9$ & $17.1$ & $819.7$ & $336.7$ & $4563.7$ & $451.8$ & $6802.1$   \\
		\textit{valid} & $1.2$ & $1.7$ & $0.13$ & $0.07$ & $3.7$ & $1.7$ & $23.52$ & $2.14$ & $34.16$ \\
		\midrule
		& Maithili & Konkani & Malayalam & Manipuri & Marathi & Nepali & Odia & Punjabi  \\
		\midrule
		\textit{train} & $113.8$ & $36.8$ & $297.7$ & $171.9$ & $458.2$ & $31.6$ & $131.4$  & $486.05$  &	$1727.45$ \\
		\textit{valid} & $0.6$ &  $0.19$ & $1.56$ &	$0.9$ &	$2.2$ & $0.18$ & $0.63$ & $2.53$	& $8.79$\\
		\midrule
		& Santali & Tamil & Telugu & Urdu & Kashmiri & Sanskrit \\
		\midrule
		\textit{train} & $6.56$ &   $542.6$ & $302.78$ & $259.68$ & $67.8$ & $58.8$& & &  $1238.22$   	\\
		\textit{valid} & $0.03$ & 	$2.67$	& $1.41$ & $1.20$ & $0.37$ & $0.30$ & & &  $5.98$	\\
		\bottomrule
	\end{tabular}
	\label{table:duration}
	\vspace*{2mm}
	\caption{Language wise duration of \textit{train} and \textit{valid} sets in hours}
\end{table}

From table $1$ it can be seen that we have a total of $9816.7$ hours of audio data out of which $9767.77$ hours is training data and $48.93$ hours is validation data in 23 Indic languages overall. To compare cross language exchange we also pretrain a model using only $4563.7$ hours of Hindi data.

\subsection{Finetuning data and Language Model}
For our current work we are using labelled data only for Hindi. The labelled data is a combination of purchased data and transcripts generated from commercial speech to text engines. We normalize the text before doing any finetuning. Any punctuation is removed and numbers are converted to word format. For language modelling we use a statistical language model based on KenLM \citep{heafield-2011-kenlm}. We use a $5$-gram language model with a beam size of $128$. The text for language model consists text in the transcribed speech and Hindi data open sourced here \footnote{\url{https://indicnlp.ai4bharat.org/corpora/}}.  

\section{Model Training}

All models are implemented in \texttt{fairseq} \citep{ott2019fairseq} library. 

\subsection{Pretraining}
We use the \textit{base} architecture of the wav2vec $2.0$ framework. It has $12$ blocks with a model dimension of $768$ and $8$ attention blocks. The pretraining is restored from a checkpoint which is trained on $960$ hours of librispeech data. We chose the base architecture over large since it is faster to train as it has almost half the parameters of large architecture. Base architecture also has a much lesser inference time when finetuned for speech recognition. We crop audio samples at $250,000$ audio frames or $15.6$ seconds and use a dropout of $0.1$. The model is trained for almost $300,000$ steps and start with a learning rate of $0.0005$. We optimize using Adam \cite{kingma2017adam} where the first $32,000$ steps are used as warmup updates for the learning rate after which it is linearly decayed. A weight of $\alpha=0.1$ is used for diversity loss $\mathcal{L}_d$ in equation \ref{total-loss}. We use $G=2$ codebooks with $V=320$ entries each for the quantization module. We train on 16 Nvidia A100 GPUs when performing pretraining on $23$ languages and on $8$ Tesla V100 GPUs when training on Hindi. It took around $100$ hours to reach a stage where we did not see any gain in code perplexity. More details about training can be seen in the training logs\footnote{\url{https://wandb.ai/harveenchadha/EKSTEP-PRETRAINING?workspace=user-agupta12}}.

\subsection{Finetuning}
To finetune a model on speech recognition downstream task, a fully connected layer is added on top of the transformer block in which the output labels are the characters for the respective language. While finetuning the weights of the feature encoder are fixed. We finetune until we get the lowest WER on the valid set. Some features of the feature encoder are masked for data augmentation. It is a technique similar to SpecAugment and detailed out in \citep{baevski2020wav2vec}. All finetuning is performed on $8$ Tesla V100 GPUs.

\begin{figure}[h]
	\centering
	\includegraphics[width=0.8\linewidth]{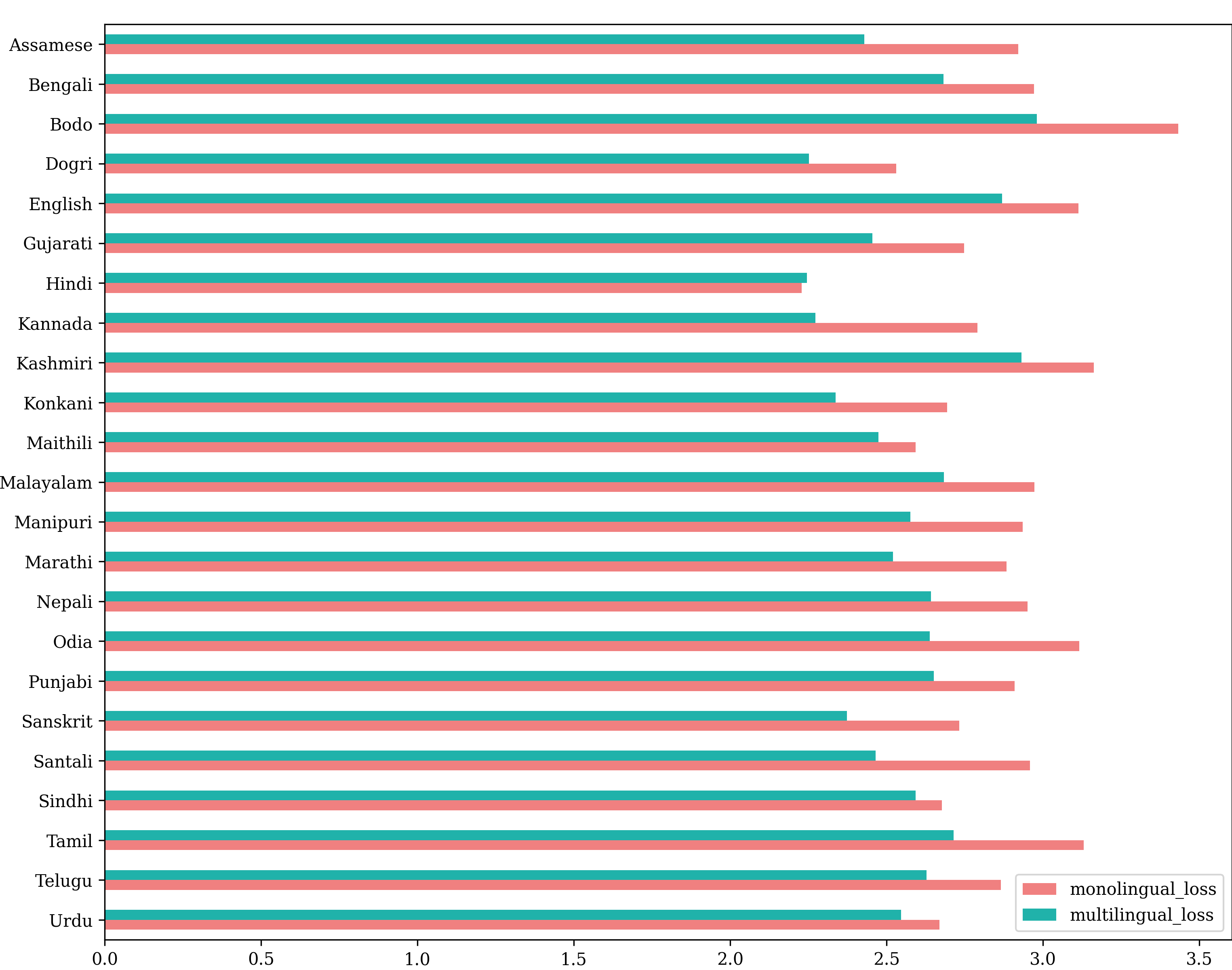}
	\caption{Language specific pretraining loss on valid set}
	\label{fig:mixedlossvalid}
\end{figure}

\section{Results}

We firstly demonstrate that multilingual pretraining outperforms monolingual pretraining by calculating language specific loss for all $23$ languages. Our experiments also show that there is a decrease in WER when multilingual pretraining model is used instead of monolingual.

\subsection{Effectiveness of Cross Lingual Representation Learning}

We calculate the language wise loss (contrastive and codebook) for audio training in both scenarios: when we have $23$ languages in pretraining and when we have just Hindi. Figure \ref{fig:mixedlossvalid} shows that for all languages apart from Hindi, the loss is lesser in the multilingual pretraining case. This is expected since low resource languages benefit from multilingual pretraining. A lower loss also indicates that more meaningful speech representations are being learnt as had been shown in \citep{conneau2020unsupervised}.

We also analyze shared discrete speech representations for different languages. For each language, we sample $200$ utterances and extract the quantized representations of the pretrained model. These vectors are normalized for each language to obtain vectors of size $ V \times G$. K-means clustering is performed on these vectors and then the dimensions are reduced by PCA. 

\begin{figure}[h]
	\centering
	\includegraphics[width=0.7\linewidth]{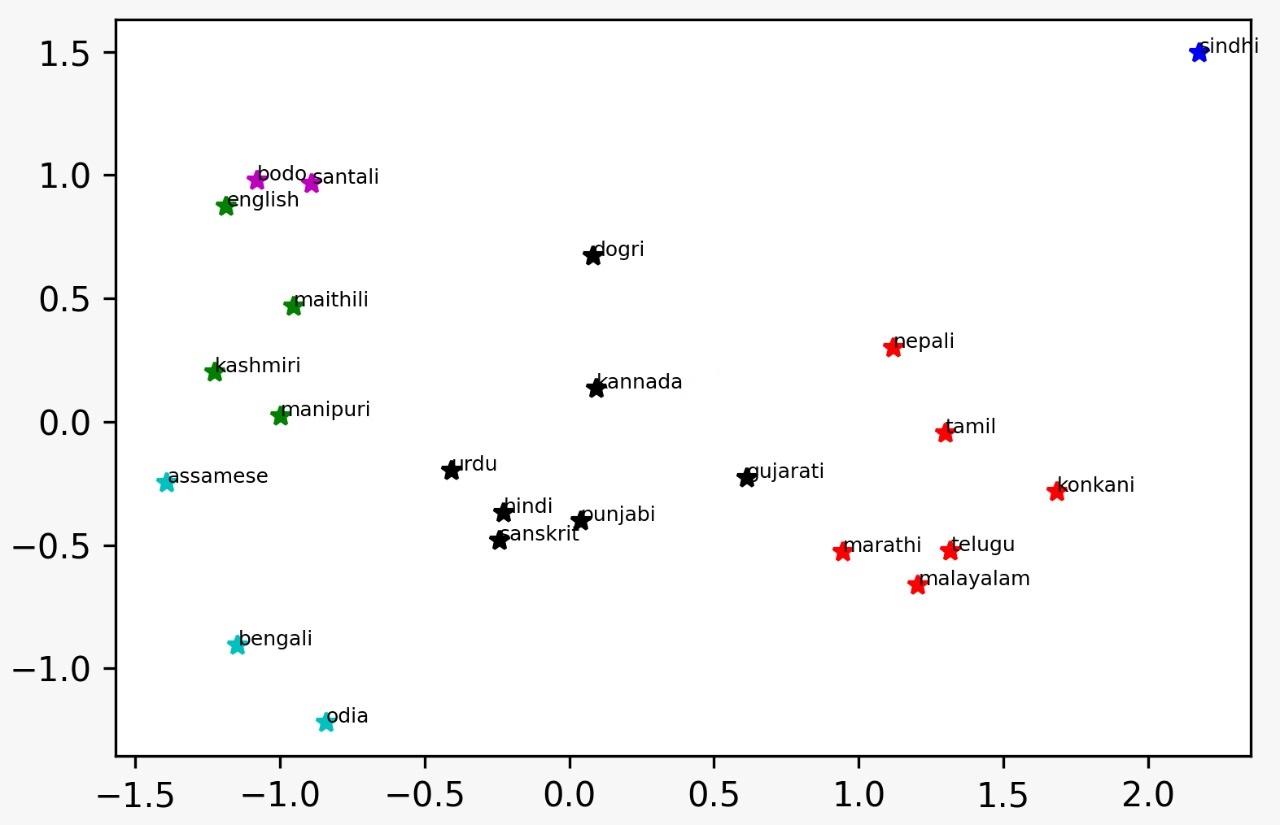}
	\caption{Quantized speech representations}
	\label{fig:quantized_speech}
\end{figure}

From \ref{fig:quantized_speech} we see that most phonetically similar languages are clustered together. The colours correspond to the clusters obtained by K-Means. We perform clustering before PCA to avoid any information loss. Assamese, Bengali and Odia are in one cluster. Hindi, Sanskrit, Urdu and Punjabi are also under one cluster. Most of the South Indian languages are clustered together as well, apart from Kannada. English in Indian accent is not a monolith. In our training data, English contains accents from different parts of India. As a result, it is far from many major languages and is together with other low resource languages. The purpose of this plot is not to recover underlying language families, but to check whether pretraining was able to learn phonetic information from a language.

\subsection{Effect on finetuning}

We finetune on $4563.7$ hours of Hindi data using checkpoints from monolingual and multilingual pretraining. We see that even the high resource language, in our case Hindi, benefits from multilingual pretraining. On a separate $10$ hour test we see a $5$\% decrease in WER and a $9.5$\% decrease in CER when decoding is done without a language model. It can be also seen from table \ref{table:wer} that the WER decrease by $3$\% and CER by $9$\% when using a language model while decoding. We also compared finetuning results on Gujarati, Tamil and Telugu using $40$ hours of training data and $5$ hours of test sets for each language. In each case we see that multilingual pretraining outperforms monolingual pretraining when compared on downatream ASR task. 

\begin{table}
	
	\centering
	\begin{tabular}{lllll}
		\toprule
		
		Pretraining  & Finetuning & Decoding & WER & CER  \\
		\midrule
		monolingual & Hindi & Viterbi & 26.2 & 9.4 \\
		multilingual & Hindi & Viterbi & 24.7 & 8.5 \\
		monolingual & Hindi & KenLM & 16.26 & 8.9 \\
		multilingual & Hindi & KenLM & 15.7 & 8.09 \\
		monolingual & Gujarati & Viterbi & 22.53 & 6.37 \\
		multilingual & Gujarati & Viterbi & 21.81 & 6.15 \\
		monolingual & Gujarati & KenLM & 16.4 & 6.2 \\
		multilingual & Gujarati & KenLM & 13.98 & 4.55 \\
		monolingual & Tamil & Viterbi & 30.08 & 6.04 \\
		multilingual & Tamil & Viterbi & 26.33 & 5.34 \\
		monolingual & Tamil & KenLM & 24.74 & 3.13 \\
		multilingual & Tamil & KenLM & 19.94 & 4.36 \\
		monolingual & Telugu & Viterbi & 29.39 & 6.2 \\
		multilingual & Telugu & Viterbi & 28.4 & 5.9 \\
		monolingual & Telugu & KenLM & 20.43 & 4.46 \\
		multilingual & Telugu & KenLM & 19.71 & 4.42 \\
		\bottomrule
		
	\end{tabular}
	\label{table:wer}
	\vspace*{2mm}
	\caption{Effect of multilingual and monolingual pretraining on WER and CER}
\end{table}

\section{Conclusion}

In this work we present a multilingual pretrained model on $23$ Indic languages in which representations are learnt from raw waveforms. Our results indicate that multilingual pretraining outperforms monolingual pretraining while learning speech representations while pre-training and also during finetuning performance in a downstream speech recognition task. We also show that model is able to encode phonetic similarity in speech representations. We hope this work kick starts development of high quality speech recognition Indic languages, especially low resource languages. All our code\footnote{\url{https://github.com/Open-Speech-EkStep/vakyansh-wav2vec2-experimentation/tree/v2-hydra}} and models\footnote{\url{https://github.com/Open-Speech-EkStep/vakyansh-models}} are open source. Our pretrained model can also be used for unsupervised speech recognition, especially for languages where no labeled text is present using the methods described in \citep{baevski2021unsupervised}. We plan to report finetuning results on all $23$ Indic languages in the future. 

\section{Acknowledgements}
All authors gratefully acknowledge Ekstep Foundation for supporting this project financially and providing infrastructure. A special thanks to Dr. Vivek Raghavan for constant support, guidance and fruitful discussions. We would also like to thank Nikita Tiwari for numerous brainstorming sessions and help in conceptualizing data pipelines and acquisition. We also thank Rajat Singhal, Heera Ballabh, Niresh Kumar R, Sreejith V, Soujyo Sen and Amulya Ahuja for automated data pipelines and infrastructure support for data processing and model training.

\bibliographystyle{unsrtnat}
\bibliography{references} 

\begin{thebibliography}{20}
\providecommand{\natexlab}[1]{#1}
\providecommand{\url}[1]{\texttt{#1}}
\expandafter\ifx\csname urlstyle\endcsname\relax
  \providecommand{\doi}[1]{doi: #1}\else
  \providecommand{\doi}{doi: \begingroup \urlstyle{rm}\Url}\fi

\bibitem[{Hinton} et~al.(2012){Hinton}, {Deng}, {Yu}, {Dahl}, {Mohamed},
  {Jaitly}, {Senior}, {Vanhoucke}, {Nguyen}, {Sainath}, and
  {Kingsbury}]{hinton}
G.~{Hinton}, L.~{Deng}, D.~{Yu}, G.~E. {Dahl}, A.~{Mohamed}, N.~{Jaitly},
  A.~{Senior}, V.~{Vanhoucke}, P.~{Nguyen}, T.~N. {Sainath}, and
  B.~{Kingsbury}.
\newblock Deep neural networks for acoustic modeling in speech recognition: The
  shared views of four research groups.
\newblock \emph{IEEE Signal Processing Magazine}, 29\penalty0 (6):\penalty0
  82--97, 2012.
\newblock \doi{10.1109/MSP.2012.2205597}.

\bibitem[Besacier et~al.(2014)Besacier, Barnard, Karpov, and
  Schultz]{besacier2014automatic}
Laurent Besacier, Etienne Barnard, Alexey Karpov, and Tanja Schultz.
\newblock Automatic speech recognition for under-resourced languages: A survey.
\newblock \emph{Speech communication}, 56:\penalty0 85--100, 2014.

\bibitem[Bourlard et~al.(2011)Bourlard, Dines, Magimai-Doss, Garner, Imseng,
  Motlicek, Liang, Saheer, and Valente]{bourlard2011current}
Herve Bourlard, John Dines, Mathew Magimai-Doss, Philip~N Garner, David Imseng,
  Petr Motlicek, Hui Liang, Lakshmi Saheer, and Fabio Valente.
\newblock Current trends in multilingual speech processing.
\newblock \emph{Sadhana}, 36\penalty0 (5):\penalty0 885--915, 2011.

\bibitem[Devlin et~al.(2019)Devlin, Chang, Lee, and Toutanova]{devlin2019bert}
Jacob Devlin, Ming-Wei Chang, Kenton Lee, and Kristina Toutanova.
\newblock Bert: Pre-training of deep bidirectional transformers for language
  understanding, 2019.

\bibitem[Peters et~al.(2018)Peters, Neumann, Iyyer, Gardner, Clark, Lee, and
  Zettlemoyer]{peters-etal-2018-deep}
Matthew~E. Peters, Mark Neumann, Mohit Iyyer, Matt Gardner, Christopher Clark,
  Kenton Lee, and Luke Zettlemoyer.
\newblock Deep contextualized word representations.
\newblock In \emph{Proceedings of the 2018 Conference of the North {A}merican
  Chapter of the Association for Computational Linguistics: Human Language
  Technologies, Volume 1 (Long Papers)}, pages 2227--2237, New Orleans,
  Louisiana, June 2018. Association for Computational Linguistics.
\newblock \doi{10.18653/v1/N18-1202}.
\newblock URL \url{https://aclanthology.org/N18-1202}.

\bibitem[Jiang et~al.(2019)Jiang, Lei, Li, Luo, Hu, Zou, and
  Li]{jiang2019improving}
Dongwei Jiang, Xiaoning Lei, Wubo Li, Ne~Luo, Yuxuan Hu, Wei Zou, and Xiangang
  Li.
\newblock Improving transformer-based speech recognition using unsupervised
  pre-training, 2019.

\bibitem[Chung and Glass(2018)]{chung2018speech2vec}
Yu-An Chung and James Glass.
\newblock Speech2vec: A sequence-to-sequence framework for learning word
  embeddings from speech.
\newblock \emph{arXiv preprint arXiv:1803.08976}, 2018.

\bibitem[Tjandra et~al.(2019)Tjandra, Sisman, Zhang, Sakti, Li, and
  Nakamura]{tjandra2019vqvae}
Andros Tjandra, Berrak Sisman, Mingyang Zhang, Sakriani Sakti, Haizhou Li, and
  Satoshi Nakamura.
\newblock Vqvae unsupervised unit discovery and multi-scale code2spec inverter
  for zerospeech challenge 2019, 2019.

\bibitem[Harwath et~al.(2020)Harwath, Hsu, and Glass]{harwath2020learning}
David Harwath, Wei-Ning Hsu, and James Glass.
\newblock Learning hierarchical discrete linguistic units from
  visually-grounded speech, 2020.

\bibitem[Baevski et~al.(2020)Baevski, Zhou, Mohamed, and
  Auli]{baevski2020wav2vec}
Alexei Baevski, Henry Zhou, Abdelrahman Mohamed, and Michael Auli.
\newblock wav2vec 2.0: A framework for self-supervised learning of speech
  representations.
\newblock \emph{arXiv preprint arXiv:2006.11477}, 2020.

\bibitem[Conneau et~al.(2020)Conneau, Baevski, Collobert, Mohamed, and
  Auli]{conneau2020unsupervised}
Alexis Conneau, Alexei Baevski, Ronan Collobert, Abdelrahman Mohamed, and
  Michael Auli.
\newblock Unsupervised cross-lingual representation learning for speech
  recognition.
\newblock \emph{arXiv preprint arXiv:2006.13979}, 2020.

\bibitem[Rivière et~al.(2020)Rivière, Joulin, Mazaré, and
  Dupoux]{rivire2020unsupervised}
Morgane Rivière, Armand Joulin, Pierre-Emmanuel Mazaré, and Emmanuel Dupoux.
\newblock Unsupervised pretraining transfers well across languages, 2020.

\bibitem[Jang et~al.(2017)Jang, Gu, and Poole]{jang2017categorical}
Eric Jang, Shixiang Gu, and Ben Poole.
\newblock Categorical reparameterization with gumbel-softmax, 2017.

\bibitem[Graves et~al.(2006)Graves, Fernández, and
  Gomez]{Graves06connectionisttemporal}
Alex Graves, Santiago Fernández, and Faustino Gomez.
\newblock Connectionist temporal classification: Labelling unsegmented sequence
  data with recurrent neural networks.
\newblock In \emph{In Proceedings of the International Conference on Machine
  Learning, ICML 2006}, pages 369--376, 2006.

\bibitem[Kim and Stern(2008)]{wadasnr}
Chanwoo Kim and Richard Stern.
\newblock Robust signal-to-noise ratio estimation based on waveform amplitude
  distribution analysis.
\newblock pages 2598--2601, 01 2008.

\bibitem[Wan et~al.(2020)Wan, Wang, Papir, and Moreno]{wan2020generalized}
Li~Wan, Quan Wang, Alan Papir, and Ignacio~Lopez Moreno.
\newblock Generalized end-to-end loss for speaker verification, 2020.

\bibitem[Heafield(2011)]{heafield-2011-kenlm}
Kenneth Heafield.
\newblock {K}en{LM}: Faster and smaller language model queries.
\newblock In \emph{Proceedings of the Sixth Workshop on Statistical Machine
  Translation}, pages 187--197, Edinburgh, Scotland, July 2011. Association for
  Computational Linguistics.
\newblock URL \url{https://www.aclweb.org/anthology/W11-2123}.

\bibitem[Ott et~al.(2019)Ott, Edunov, Baevski, Fan, Gross, Ng, Grangier, and
  Auli]{ott2019fairseq}
Myle Ott, Sergey Edunov, Alexei Baevski, Angela Fan, Sam Gross, Nathan Ng,
  David Grangier, and Michael Auli.
\newblock fairseq: A fast, extensible toolkit for sequence modeling, 2019.

\bibitem[Kingma and Ba(2017)]{kingma2017adam}
Diederik~P. Kingma and Jimmy Ba.
\newblock Adam: A method for stochastic optimization, 2017.

\bibitem[Baevski et~al.(2021)Baevski, Hsu, Conneau, and
  Auli]{baevski2021unsupervised}
Alexei Baevski, Wei-Ning Hsu, Alexis Conneau, and Michael Auli.
\newblock Unsupervised speech recognition, 2021.

\end{thebibliography}

\end{document}